\documentclass{article}

\usepackage[nonatbib, preprint]{neurips_2018}
\usepackage[utf8]{inputenc} 
\usepackage[T1]{fontenc}    
\usepackage{hyperref}       
\usepackage{url}            
\usepackage{booktabs}       
\usepackage{amsfonts}       
\usepackage{nicefrac}       
\usepackage{microtype}      
\usepackage{lipsum}
\usepackage{graphicx}
\usepackage[ruled,vlined]{algorithm2e}

\newcommand{\N}{\mathbb{N}}

\usepackage{amsmath}
\usepackage{tabularx, booktabs, caption, makecell, float}
\usepackage{subfig}
  \usepackage{multirow}

\title{Network Generation with Differential Privacy}

\author{
  Xu Zheng \\
  Accenture Labs\\
  Dublin, Ireland\\
  \texttt{xu.b.zheng@accenture.com} \\
  \And
  Nicholas McCarthy \\
  Accenture Labs\\
  Dublin, Ireland\\
  \texttt{nicholas.mccarthy@accenture.com} \\
  \And
  Jer Hayes \\
  Accenture Labs\\
  Dublin, Ireland\\
  \texttt{jeremiah.hayes@accenture.com} \\
}

\begin{document}

\maketitle
\begin{abstract}
We consider the problem of generating private synthetic versions of real-world graphs containing private information while maintaining the utility of generated graphs. Differential privacy is a gold standard for data privacy, and the introduction of the differentially private stochastic gradient descent (DP-SGD) algorithm has facilitated the training of private neural models in a number of domains. Recent advances in graph generation via deep generative networks have produced several high performing models. We evaluate and compare state-of-the-art models including adjacency matrix based models and edge based models, and show a practical implementation that favours the edge-list approach utilizing the Gaussian noise mechanism, when evaluated on commonly used graph datasets. Based on our findings, we propose a generative model that can reproduce the properties of real-world networks while maintaining edge-differential privacy. The proposed model is based on a stochastic neural network that generates discrete edge-list samples and is trained using the Wasserstein GAN objective with the DP-SGD optimizer. Being the first approach to combine these beneficial properties, our model contributes to further research on graph data privacy.
\end{abstract}





\section{Introduction}

Advances in generative models for graphs driven by the rise of deep learning have led to a number of proposed applications including data imputation, novel molecule generation, and knowledge discovery \cite{wang2018graphgan,decao2018molgan}. Several works have used generative adversarial network (GAN) architectures \cite{goodfellow2014generative} in the problem of generating synthetic versions of real-world graphs by operating directly on adjacency matrices \cite{tavakoli2017learning,fan2019labeled}. However, this approach has a number of problems: foremost it has limited scalability as the computation and memory requirements scale quadratically with the number of graph vertices, making it feasible to process only small graphs. Moreover, they have mixed results when attempting to generate graphs that match the statistical properties of real-world graphs \cite{Dong2017}. Node permutation can be used to create different variants of the graph in order to train a GAN, but this can significantly affect graph structure resulting in difficulties training the generator, requires node ordering heuristics or other graph matching operations. 



Node embedding methods achieve state-of-the-art scores in tasks like link prediction and node classification \cite{wang2018graphgan,perozzi2014deepwalk,grover2016node2vec}. The main idea behind these approaches is to model the probabilities of each edge’s existence by a neural network. Nevertheless, such approaches still have difficulty in preserving and generating patterns inherent to real-world networks. The NetGAN model \cite{bojchevski2018netgan} applies random walks on the input graphs \cite{lovasz1993random}, and an LSTM-based generator tries to mimic these walk sequences \cite{hochreiter1997long}. As it considers only the non-zero entries of the adjacency matrix it avoids the heavy computational requirements for operating on adjacency matrices, and is readily applicable to graphs with thousands of nodes. 

Although edge-based GAN models can synthesize graphs that closely model the original data distribution, the existence of private data in the graph is not taken into account, and the generated graph can breach individual privacy by leaking personally identifiable information. Among research communities and in certain domains such as medical, healthcare and insurance there is a significant advantage to generating synthetic representations of private data. However, synthetic data generated using standard generative model architectures is not private \cite{jordon2018pate}. Privacy-preserving release of tabular and image data \cite{augenstein2019generative} has been shown to be possible by implementing a generative adversarial framework with differential privacy mechanisms \cite{dwork2018privacy, xie2018differentially}. Hence, integrating generative models and differential privacy mechanisms for generating graphs that can maintain both statistical and topological information of the original graphs opens new avenues for further research. 
The contributions of our work are as follows. Firstly we propose a new model for generating graphs using an edge list. This model is able to learn from a single graph and generate discrete outputs that maintain topological properties of the original graph. Secondly, by engaging the edge list we successfully integrated differential privacy mechanisms into the training procedure. Theoretically, we show that our algorithm ensures differential privacy for the generated graphs, and demonstrate empirically its performance in experiments on two datasets. The results show that our method is competitive with baseline model NetGAN \cite{bojchevski2018netgan} in terms of both statistical properties and data utility. 

\section{Method}

\begin{figure}[h]
\centering
\includegraphics[width=0.98\linewidth]{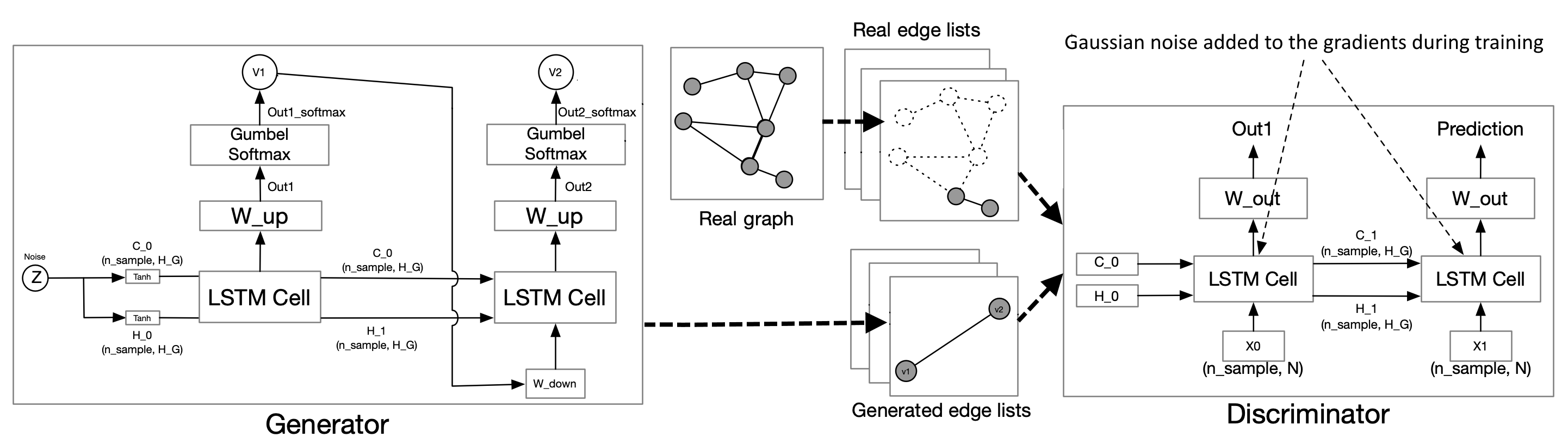}
\caption{The architecture of DP-NetGAN. The Generator and Discriminator are based on an LSTM architecture, and the inputs to the model are the edge lists of the graph. Calibrated noise is added to the discriminator during training.}
\label{xgan_architecture}
\end{figure}

In this section we present our model for synthesizing graphs with edge-differential privacy. The overall architecture is based on NetGAN \cite{bojchevski2018netgan}, which operates random walks on the input graph \cite{lovasz1993random}. We modify the training procedure by replacing the random walk method and instead represent the input graph $G$ by a sparse adjacency tuple $I$, where $I \in \N^{2} $ encodes edge indices in COOrdinate format \cite{Fey/Lenssen/2019}. By modifying and using an edge list instead of random walk, we can train the generative model with edge-differential privacy. As with typical GAN architectures, this model contains two main components: a generator $g$, which is trained to generate synthetic edge lists, and a discriminator $d$ that learns to distinguish the synthetic edge lists from the real ones that are sampled from the real set of edges. We train our model based on the Wasserstein GAN (WGAN) framework because it is more stable and able to prevent mode collapse \cite{arjovsky2017wasserstein}. The architecture can found in Figure \ref{xgan_architecture}, and the gradient perturbation algorithm can be found in algorithm \ref{appendix:algo_dp}. 

\subsection{Differential Privacy}
Differential privacy can be achieved by injecting Gaussian noise parameterized by sensitivity to the clipped gradients in the optimization procedure when training $d$. In this work we use the differentially private stochastic gradient descent (DP-SGD) algorithm \ref{appendix:algo_dp} to sanitize the gradients of discriminator $d$ and add calibrated Gaussian noise. The noise is scaled by the noise factor $\sigma$ defined in algorithm \ref{appendix:algo_dp} and the clip value $C$ applied on the gradients represents the sensitivity in differential privacy mechanisms. A key component of the model is to keep track of the cumulative privacy loss during the training phase. Due to the composability property of differential privacy, each training step's privacy cost can be accumulated by a privacy accountant. We employ the Moments Accounting procedure proposed in \cite{abadi2016deep} that additively accumulates the log of the moments of the privacy loss at each step. This method provides a tighter estimation of the privacy loss and allows us to compute the overall privacy costs at each iteration.

\subsection{Assembling Graph from Edge List}\label{graph_generation}

\begin{figure}[h]
\centering
\includegraphics[width=0.7\linewidth]{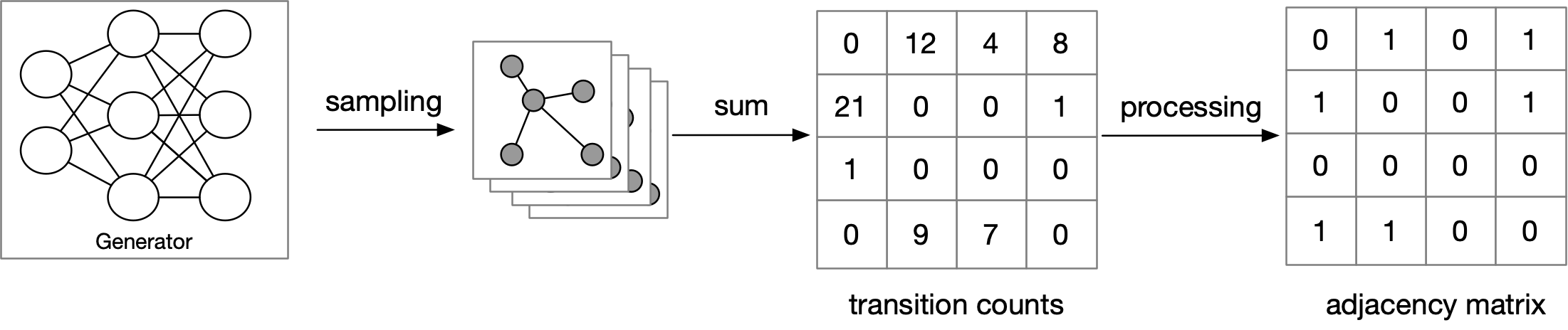}
\caption{Constructing an adjacency matrix from raw edge lists output by the generator. }
\label{matrix_assembling}
\end{figure}

Following these steps we can use an edge list to compose an adjacency matrix that represents a synthetic graph, illustrated in Figure \ref{matrix_assembling}. We firstly use the trained generator $g$ to generate a large number of edge lists. In work \cite{bojchevski2018netgan} it has been shown that a larger number here results in better graphs. A count of how often an edge appears in the set of generated edge lists is used to create a square matrix $S$. In this matrix we can not guarantee the symmetry property in the input graph as it is undirected, so we set $s_{ij} = s_{ji} = max(s_{ij}, s_{ji})$ explicitly. If we just apply a threshold on the matrix to construct an adjacency matrix, nodes with a high degree tend to be over-represented and low degree nodes can be left out. Hence, we firstly ensure every node $i$ has at least one edge by sampling a neighbor $j$ with probability $p_{ij} = \frac{s_{ij}}{\sum_{v} S{iv}}$. We continue sampling for each edge $(i, j)$ with the probability $p_{i,j} = \frac{s_{ij}}{\sum_{u,v} S{uv}}$ until we reach as many unique edges as in the original graph. For each $(i, j)$ we also include $(j, i)$ so that we can have an undirected graph. We can then convert this raw counts matrix to a binary adjacency matrix. 

\section{Experiments and Results}

We evaluate the quality of the generated graphs from two aspects: graph statistical properties and graph utility. The graph statistical properties include the graph maximum degrees, assortativity, triangle count, the power-law exponent, clustering coefficient and characteristic path length (definitions in Appendix \ref{graph_statistics}). Graph utility is evaluated based on link prediction performance on the validation set. We chose two well-known datasets for our experiments, CORA-ML \cite{mccallum2000automating} and Citeseer \cite{sen2008collective}. For these experiments we treat the graphs as undirected and only consider the largest connected component (LCC) so that we can compare the results with the baseline NetGAN model \cite{bojchevski2018netgan}, which requires the graph to be connected. We randomly sample 15\% of the edges as a validation set, and fit two models to the remaining training set. We trained NetGAN with the standard 16-step random walk and a 2-step random walk for comparison with the edge list approach. Early stopping is used to stop the model training if performance on the validation set does not increase in previous 5 checkpoints As the evaluation process is time consuming and resource consuming, each checkpoint contains $m$ epochs, while $m$ is a parameter set before training.



\subsection{Graph Statistics} In our experiments we set the \textit{batch size} = 2048 and $\delta$ = 10e-5 for both datasets. Gradient clip values are selected from \{0.05, 0.1, 0.5, 1\}, as we observed that most of the gradients of the discriminator are smaller than 1 for the standard NetGAN model. The value of $\epsilon$ is then computed for different noise levels and training epochs $E$ with the moments accountant. For both DP-NetGAN and NetGAN, we generate 400k edge lists or walks and construct the adjacency matrix as the final generated graph at each checkpoint. Graph statistics are calculated on these generated graphs with different level of epsilon values. In Table \ref{table:1}, we observe that DP-NetGAN captures the graph properties well, and note that the increasing epsilon values result in a relatively stable improvement in graph statistics: the privacy-utility trade-off. It's worth noting that NetGAN with 16-steps achieves highest performance over all metrics, but that DP-NetGAN with $\epsilon=\inf$ surpasses NetGAN with 2 steps due to the change in training protocol. This is most obvious the \#Triangles statistic as the 2-step random walk fails to capture local triangular patterns of the input graph.  


To increase edge privacy, we can increase the noise added during training, resulting in smaller epsilon values. Although the edge overlap between input and generated graphs is relatively small (<20\%) for DP-NetGAN, the graph statistics are comparable, implying that DP-NetGAN does not just memorize a subset of edges and randomly generate the remainder, but instead captures the underlying structure of the network.

\begin{table}[h!]
\centering
\caption{Table of graph statistics on the output of models at different epsilon values.}
\begin{tabular}{ |c||c|c|c|c|c|c|}
 \hline
 \textbf{Model / Data} & \textbf{MaxDeg}.& \textbf{Assor.} & \textbf{\#Triangle} & \textbf{PLE} & \textbf{Cluster} & \textbf{CPL} \\ 
 \hline \hline \hline
\textbf{CORA-ML} & \textbf{ 240} & \textbf{-0.075} & \textbf{2814} & \textbf{1.86} & \textbf{2.7e-3} & \textbf{5.61}\\ \hline \hline
 DP-NetGAN($\epsilon$=2.9)&200&-0.258&386&1.84&2e-4&3.81\\ \hline
 DP-NetGAN($\epsilon$=9.5)&138&-0.134&393&1.86&7.6e-4&4.00\\ \hline
 DP-NetGAN($\epsilon$=440)&249&-0.036&829&1.72&6.7e-4&4.32\\ \hline
 DP-NetGAN($\epsilon$=inf)&220&-0.058&807&1.72&1e-3&4.33\\ \hline
 NetGAN(2steps)&151&-0.101&249&1.68&7.5e-4&4.40\\  \hline
 NetGAN(16steps)&262&-0.065&1830&1.71&1.4e-3&4.48\\ \hline
 \hline \hline
 \textbf{CITESEER} & \textbf{99} & \textbf{0.0075} & \textbf{1084} & \textbf{2.07} & \textbf{1.3e-2} & \textbf{9.33}\\ \hline \hline
 DP-NetGAN($\epsilon$=0.68)&400&-0.212&434&2.28&7e-4&4.9\\ \hline
 DP-NetGAN($\epsilon$=16)&276&-0.075 & 78 & 2.18 & 6.35e-5&4.69\\ \hline
 DP-NetGAN($\epsilon$=296)&74&-0.052&26&2.03&6.34e-4&5.5\\ \hline
 DP-NetGAN($\epsilon$=inf)&54&-0.024&155&1.96&7.3e-3&6.28\\ \hline
 NetGAN(2steps)&32&-0.167&54& 1.96 & 5.3e-3&6.19\\ \hline
 NetGAN(16steps)&55&-0.025&493 &2.02 & 1.7e-2&6.88\\ \hline
\end{tabular}
\label{table:1}
\end{table}

\subsection{Link Prediction.} Link prediction is a standard task in graph learning where the goal is to estimate the probability of links between nodes in a graph. We use the area under the ROC curve (AUC) and average precision (AP) to evaluate the link prediction performance on the validation set. We can see from table \ref{table:2} that our model shows competitive performance for two datasets compared with NetGAN 2-steps, and our model's performance increases when we increase the $\epsilon$ values.

\begin{table}[h!]
\centering
\caption{Link prediction performance on validation set.}
\begin{tabular}{ |p{3.5cm}||p{1.5cm}|p{1.5cm}|p{1.5cm}|p{1.5cm}|}
\hline
~&\multicolumn{2}{|c|}{Cora-ML} & \multicolumn{2}{|c|}{Citeseer}\\
 \hline
Model & AUC & AP & AUC & AP\\
 \hline
 DP-NetGAN ($\epsilon$=*1e-1)&60.12&60.07&69.68&68.73\\\hline
 DP-NetGAN ($\epsilon$=*1e2)&80.04&80.49&72.03&75.42\\\hline
 DP-NetGAN ($\epsilon$=*1e3)&89.29&90.73&89.29&90.73\\\hline
 DP-NetGAN ($\epsilon$=inf)&88.44&89.9&83.47&86.18\\\hline
 NetGAN (2steps)&85.33&86.72&80.88&83.76\\\hline
 NetGAN (16steps)&96.17&96.65&95.33&96.38\\\hline
\end{tabular}
\label{table:2}
\end{table}
\section{Discussion and Future Work}
In this work we introduced a scalable differentially private generative model for graph data. By utilizing edge lists, this model can generate networks that capture essential topological properties. It also shows competitive link prediction performance while maintaining edge-differential privacy. Future research directions could look at private synthesis of multi-relational graph data such as knowledge graphs, which could be achieved by a relatively simple modification of the architecture proposed here by introducing a relation step into the LSTM input, and embedding layers for node and edge to maintain corresponding input dimensions. Alternatively, examining differential privacy mechanisms such as PATE in a graph generation context would also be of use. 



\bibliographystyle{unsrt}  
\bibliography{references}

\begin{thebibliography}{10}

\bibitem{wang2018graphgan}
Hongwei Wang, Jia Wang, Jialin Wang, Miao Zhao, Weinan Zhang, Fuzheng Zhang,
  Xing Xie, and Minyi Guo.
\newblock Graphgan: Graph representation learning with generative adversarial
  nets.
\newblock In {\em Thirty-second AAAI conference on artificial intelligence},
  2018.

\bibitem{decao2018molgan}
Nicola~De Cao and Thomas Kipf.
\newblock Molgan: An implicit generative model for small molecular graphs,
  2018.

\bibitem{goodfellow2014generative}
Ian Goodfellow, Jean Pouget-Abadie, Mehdi Mirza, Bing Xu, David Warde-Farley,
  Sherjil Ozair, Aaron Courville, and Yoshua Bengio.
\newblock Generative adversarial nets.
\newblock In {\em Advances in neural information processing systems}, pages
  2672--2680, 2014.

\bibitem{tavakoli2017learning}
Sahar Tavakoli, Alireza Hajibagheri, and Gita Sukthankar.
\newblock Learning social graph topologies using generative adversarial neural
  networks.
\newblock In {\em International Conference on Social Computing,
  Behavioral-Cultural Modeling \& Prediction}, 2017.

\bibitem{fan2019labeled}
Shuangfei Fan and Bert Huang.
\newblock Labeled graph generative adversarial networks.
\newblock {\em CoRR}, abs/1906.03220, 2019.

\bibitem{Dong2017}
Yuxiao Dong, Reid~A. Johnson, Jian Xu, and Nitesh~V. Chawla.
\newblock {Structural diversity and homophily: A study across more than one
  hundred big networks}.
\newblock {\em Proceedings of the ACM SIGKDD International Conference on
  Knowledge Discovery and Data Mining}, Part F129685:807--816, 2017.

\bibitem{perozzi2014deepwalk}
Bryan Perozzi, Rami Al-Rfou, and Steven Skiena.
\newblock Deepwalk: Online learning of social representations.
\newblock In {\em Proceedings of the 20th ACM SIGKDD international conference
  on Knowledge discovery and data mining}, pages 701--710, 2014.

\bibitem{grover2016node2vec}
Aditya Grover and Jure Leskovec.
\newblock node2vec: Scalable feature learning for networks.
\newblock In {\em Proceedings of the 22nd ACM SIGKDD international conference
  on Knowledge discovery and data mining}, pages 855--864, 2016.

\bibitem{bojchevski2018netgan}
Aleksandar Bojchevski, Oleksandr Shchur, Daniel Z{\"{u}}gner, and Stephan
  G{\"{u}}nnemann.
\newblock Netgan: Generating graphs via random walks.
\newblock In {\em Proceedings of the 35th International Conference on Machine
  Learning, {ICML} 2018, Stockholmsm{\"{a}}ssan, Stockholm, Sweden, July 10-15,
  2018}, pages 609--618, 2018.

\bibitem{lovasz1993random}
L{\'a}szl{\'o} Lov{\'a}sz et~al.
\newblock Random walks on graphs: A survey.
\newblock {\em Combinatorics, Paul erdos is eighty}, 2(1):1--46, 1993.

\bibitem{hochreiter1997long}
Sepp Hochreiter and J{\"u}rgen Schmidhuber.
\newblock Long short-term memory.
\newblock {\em Neural computation}, 9(8):1735--1780, 1997.

\bibitem{jordon2018pate}
James Jordon, Jinsung Yoon, and Mihaela Van Der~Schaar.
\newblock Pate-gan: Generating synthetic data with differential privacy
  guarantees.
\newblock In {\em International conference on learning representations}, 2018.

\bibitem{augenstein2019generative}
Sean Augenstein, H~Brendan McMahan, Daniel Ramage, Swaroop Ramaswamy, Peter
  Kairouz, Mingqing Chen, Rajiv Mathews, et~al.
\newblock Generative models for effective ml on private, decentralized
  datasets.
\newblock {\em ICLR}, 2020.

\bibitem{dwork2018privacy}
Cynthia Dwork and Vitaly Feldman.
\newblock Privacy-preserving prediction.
\newblock {\em Conference on Learning Theory (COLT)}, 2018.

\bibitem{xie2018differentially}
Liyang Xie, Kaixiang Lin, Shu Wang, Fei Wang, and Jiayu Zhou.
\newblock Differentially private generative adversarial network.
\newblock {\em arXiv preprint arXiv:1802.06739}, 2018.

\bibitem{Fey/Lenssen/2019}
Matthias Fey and Jan~E. Lenssen.
\newblock Fast graph representation learning with {PyTorch Geometric}.
\newblock In {\em ICLR Workshop on Representation Learning on Graphs and
  Manifolds}, 2019.

\bibitem{arjovsky2017wasserstein}
Martin Arjovsky, Soumith Chintala, and L{\'e}on Bottou.
\newblock Wasserstein generative adversarial networks.
\newblock In {\em Proceedings of the 34th International Conference on Machine
  Learning-Volume 70}, pages 214--223, 2017.

\bibitem{abadi2016deep}
Martin Abadi, Andy Chu, Ian Goodfellow, H~Brendan McMahan, Ilya Mironov, Kunal
  Talwar, and Li~Zhang.
\newblock Deep learning with differential privacy.
\newblock In {\em Proceedings of the 2016 ACM SIGSAC Conference on Computer and
  Communications Security}, pages 308--318, 2016.

\bibitem{mccallum2000automating}
Andrew~Kachites McCallum, Kamal Nigam, Jason Rennie, and Kristie Seymore.
\newblock Automating the construction of internet portals with machine
  learning.
\newblock {\em Information Retrieval}, 3(2):127--163, 2000.

\bibitem{sen2008collective}
Prithviraj Sen, Galileo Namata, Mustafa Bilgic, Lise Getoor, Brian Galligher,
  and Tina Eliassi-Rad.
\newblock Collective classification in network data.
\newblock {\em AI magazine}, 29(3):93--93, 2008.

\end{thebibliography}

\appendix


\clearpage

\section{Graph Statistics}
\begin{table}[htb]
  \begin{minipage}{0.5\textheight}
  \caption{Graph Statistics List}
  \label{graph_statistics}  
  \begin{tabular}{l*{2}{l}}
    Metric name& Description \\
    \midrule
    Max Degree& Maximum degree of all nodes in a graph\\
    Assortativity& Pearson correlation of degrees of connected nodes\\
    Triangle Count& Number of triangles in the graph\\
    Power law exponent& Exponent of the power law distribution\\
    Clustering coeff& The degree to which nodes tend to cluster together\\
    Characteristic path len&  Average steps along the shortest paths for all pairs\\
  \end{tabular}
  \end{minipage}
\end{table}


\section{DP GAN}
\begin{figure}[htb]
\centering
\begin{minipage}{.9\linewidth} 
\IncMargin{1em}
\begin{algorithm}[H]
\SetKwData{Left}{left}\SetKwData{This}{this}\SetKwData{Up}{up}
\SetKwFunction{Union}{Union}\SetKwFunction{Find Compress}{Find Compress}
\SetKwInOut{Input}{input}\SetKwInOut{Output}{output}
\Input{$n$ - number of samples;  $\lambda$ - coefficient of gradient penalty; $n_{critic}$ - number of critic iterations per generator iteration; $n_{param}$ - number of discriminator's parameters; $m$ - batch size; $(\alpha, \beta_1, \beta_2)$ - Adam hyper-parameters; $C$ - gradient clipping bound; $\sigma$ - noise scale;}
\Output{A differentially private generator $G$}
\While{$\theta$ has not converged}{
    \For{$i\leftarrow 1$ \KwTo $n_{critic}$}{
        \For{$j\leftarrow 1$ \KwTo $m$}{
            sample $x \sim p_{data}, z \sim p_z, \rho \sim \mu[0,1]$ \;
            $\hat{x} \leftarrow px + (1 - p)G(z)$ \;
            $l^{(i)} \leftarrow D(G(z)) - D(x) + \lambda(\Vert\Delta_{\hat{x}}D(\hat{x})\Vert_{2} - 1)^2$ \;
            $g^{(i)} \leftarrow \Delta_{w}l^{(i)}$ \; 
            $g^{(i)} \leftarrow g^{(i)}/max(1, \Vert g^{(i)} \Vert_{2}/C)$ \;
        }
        // Gaussian noise \;
        $\xi \sim N(0, (\sigma C)^{2}I) $ \;
        $\hat{g} = \frac{1}{m}(\sum_{i=1}^{m} g^{(i)} + \xi $)\;
        $w \leftarrow Adam(\hat{g}, w, \alpha, \beta_{1}, \beta_{2})$ \;
        update A with $(\theta,m,n_{param})$ \;
    }
    sample $\{z^{(i)}\}_{i=1}^{m} \sim p_{z}$ \;
    // updating generator \;
    $\theta \leftarrow Adam(\Delta_{\theta} \frac{1}{m} \sum_{i=1}^{m} -D(G(z^{(i)})), \theta, \alpha, \beta_{1}, \beta_{2})$ \;
    // computing cumulative privacy loss \;
    $\delta \leftarrow$ query A with $\epsilon_{0}$ \;
    \If{$\delta > \delta_{0}$}{ break}
}
\BlankLine
\Return{$G$}
\caption{Basic DP GAN}
\label{appendix:algo_dp}
\end{algorithm} \DecMargin{1em}
\end{minipage}
\end{figure}




\end{document}